\documentclass{article} 
\usepackage{iclr2025_conference,times}

\usepackage{amsmath,amsfonts,bm}

\def\eqref#1{equation~\ref{#1}}

\def\1{\bm{1}}

\def\vc{{\bm{c}}}

\DeclareMathAlphabet{\mathsfit}{\encodingdefault}{\sfdefault}{m}{sl}
\SetMathAlphabet{\mathsfit}{bold}{\encodingdefault}{\sfdefault}{bx}{n}

\usepackage{hyperref}
\usepackage{url}

\usepackage{graphicx}
\usepackage{subfigure}
\usepackage{pifont}
\usepackage{setspace} 
\usepackage{booktabs}
\usepackage{colortbl}
\usepackage{dsfont}
\usepackage{bbm}
\usepackage{multirow}
\usepackage{amsmath}
\usepackage{amssymb}
\usepackage{mathtools}
\usepackage{amsthm}
\usepackage{algorithm}
\usepackage{algpseudocode}
\usepackage{caption}
\DeclareCaptionLabelFormat{alglabel}{Algorithm~#2}
\newcommand{\eg}{\textit{e}.\textit{g}.}

\newcommand{\ie}{\textit{i}.\textit{e}.}

\definecolor{mydarkblue}{rgb}{0,0.08,0.45}
\definecolor{darkred}{rgb}{0.55,0,0}
\usepackage{hyperref}
\hypersetup{
    colorlinks=true,
    citecolor=mydarkblue, 
    linkcolor=darkred,    
    urlcolor=mydarkblue
}

\title{Modality-Composable Diffusion Policy \\ via Inference-Time Distribution-level Composition}

\def\mystrut{\rule{0pt}{1.0\normalbaselineskip}}
\author{
\begin{tabular}{@{}l}
Jiahang Cao$^{1,2}$, Qiang Zhang$^{1,2}$, Hanzhong Guo$^{3}$, Jiaxu Wang$^{1}$, Hao Cheng$^{1}$ \& Renjing Xu$^{1}$\mystrut \\
\end{tabular}\\
$^1$The Hong Kong University of Science and Technology (Guangzhou)\mystrut\quad\\
$^2$Beijing Innovation Center of Humanoid Robotics\\
$^3$The University of Hong Kong\\
~\texttt{jcao248@connect.hkust-gz.edu.cn} 
}

\iclrfinalcopy 
\begin{document}

\maketitle

\begin{abstract}
Diffusion Policy (DP) has attracted significant attention as an effective method for policy representation due to its capacity to model multi-distribution dynamics. However, current DPs are often based on a single visual modality (\eg, RGB or point cloud), limiting their accuracy and generalization potential. Although training a generalized DP capable of handling heterogeneous multimodal data would enhance performance, it entails substantial computational and data-related costs. To address these challenges, we propose a novel policy composition method: by leveraging multiple pre-trained DPs based on individual visual modalities, we can combine their distributional scores to form a more expressive Modality-Composable Diffusion Policy (MCDP), without the need for additional training. Through extensive empirical experiments on the RoboTwin dataset, we demonstrate the potential of MCDP to improve both adaptability and performance. This exploration aims to provide valuable insights into the flexible composition of existing DPs, facilitating the development of generalizable cross-modality, cross-domain, and even cross-embodiment policies. Our code is open-sourced at \url{https://github.com/AndyCao1125/MCDP}.
\end{abstract}

\begin{figure}[h]
    \centering
    \includegraphics[width = .8\linewidth]{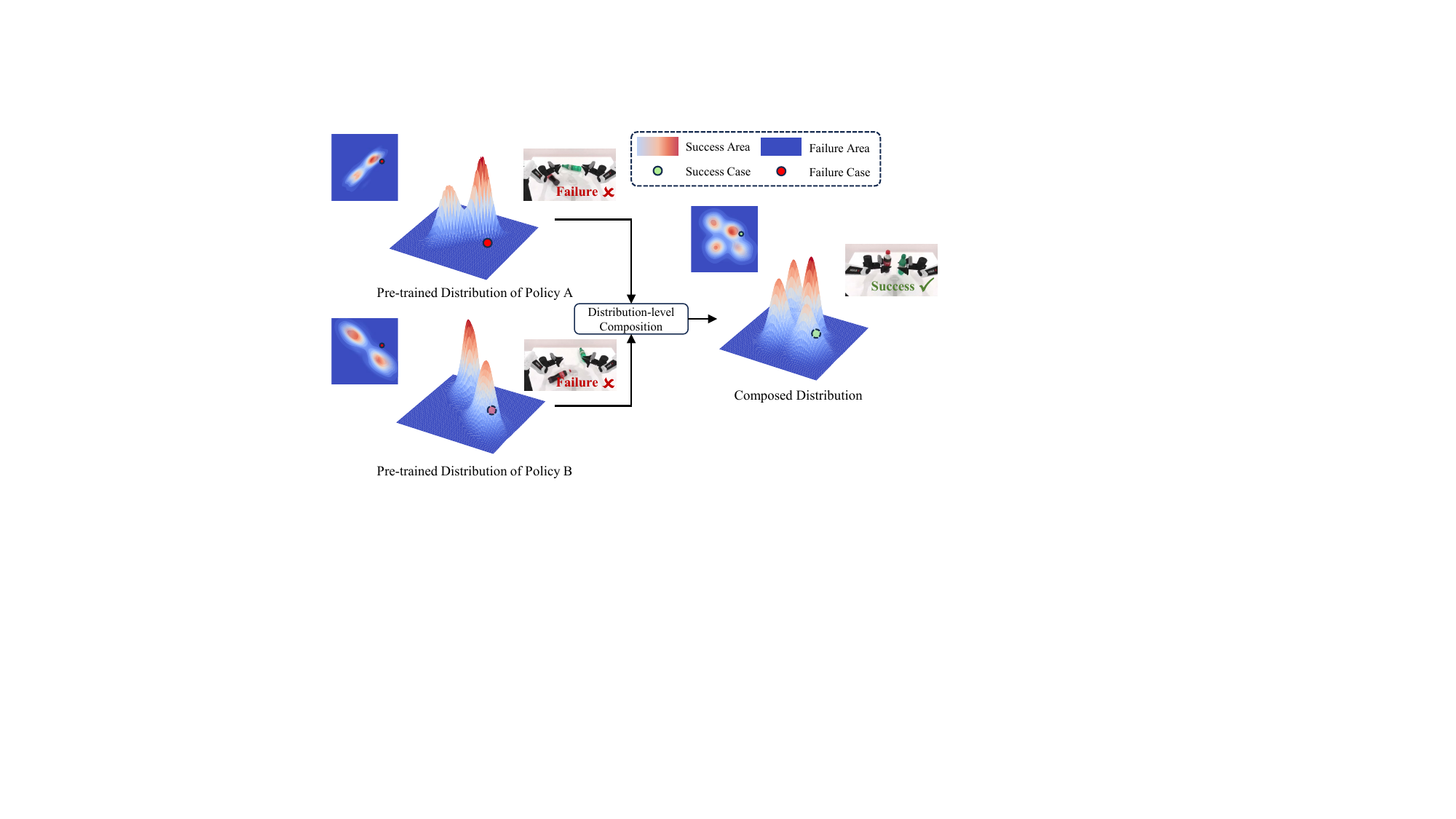}
    \caption{\textbf{Illustrative Example of Modality-Composable Diffusion Policy.} Distributions from pre-trained policies based on single-visual modalities can be combined to construct a stronger policy.}
    \label{fig:teaser}
    \vspace{-3mm}
\end{figure}

\section{Introduction}

Diffusion Policy (DP)~\citep{chi2023diffusion} has emerged as an effective method for policy representation in robot learning, owing to its distinctive ability to model multi-modal action distributions and impressive training stability. By harnessing the power of diffusion models~\citep{ho2020denoising,songdenoising}, DP provides a flexible and robust framework for complex policy learning, facilitating its adoption in a wide range of applications, including manipulation~\citep{ze20243d,zhu2024scaling} and navigation~\citep{sridhar2024nomad,zhang2024versatile}.

Despite their potential, current DP frameworks typically rely on a single visual modality, \eg, RGB~\citep{chi2023diffusion,reuss2024multimodal} or point cloud~\citep{chi2023diffusion,cao2024mamba}. This limitation prevents them from fully capturing the complexity of real-world scenes, leading to low accuracy and suboptimal decision-making. One possible solution is to collect multimodal data~\citep{wu2024robomind} and train DPs capable of handling mult-modalities~\citep{huang2025enerverse}. However, this approach comes with significant data collection costs, as learning joint distributions over multiple variables requires exponentially more samples than learning a distribution on a single variable~\citep{canonne2020short}. Another alternative is to adopt vision language models~\citep{karamchetiprismatic} to create vision language action models~\citep{kim2024openvla} for multimodal understanding. Yet, this approach is also hindered by the substantial computational resources for training.

Given these challenges, a key question arises: how can we enhance the accuracy of DP without incurring excessive costs? Compositional diffusion models~\citep{du2024compositional,wang2024poco} offer a potential solution, where the diffusion scores from different expertized models can be composed at inference time to construct a stronger generative system. 
Inspired by this, we propose a novel approach: Modality-Composable Diffusion Policy (MCDP, Fig.~\ref{fig:teaser}). By leveraging multiple pre-trained DPs, each based on a single visual modality, we can combine their diffusion scores during the sampling process to form a more expressive and capable policy, \textit{without the need to modify pre-trained policies}. Extensive experiments on RoboTwin dataset~\citep{mu2024robotwin} demonstrate that MCDP achieves superior performance than the unimodal DPs. In addition, we evaluate different weight combinations for DP composition and analyze sample distribution through visualization, providing meaningful insights for policy composition strategies.

Our contributions are summarized as follows:
\textbf{(I)} We propose Modality-Composable Diffusion Policy (MCDP), a novel framework for combining DPs from a visual modality perspective, which leads to a significant improvement without additional training. \textbf{(II)} We conduct extensive evaluations of the weight combinations used in policy composition, analyzing their impact on policy performance across different scenarios. \textbf{(III)} Based on our empirical experiments on the RoboTwin dataset, we provide valuable findings for policy composition that could be applied to the development of generalizable cross-modality, cross-domain, and even cross-embodiment policies in the future.

\section{Preliminaries}
\paragraph{Denoising Diffusion Probabilistic Model (DDPM).}
Diffusion models, including DDPM~\citep{ho2020denoising}, are based on a generative process that iteratively denoises a random noise distribution to generate samples. 
The following equation describes the update rule for the diffusion process based on Langevin dynamics~\citep{song2020score}:
\begin{equation}
    \tau^{t-1}=\alpha^t(\tau^{t}-\gamma^t \epsilon_\theta(\tau^t,t) +\xi),\quad \xi\sim \mathcal{N} \bigl(\textbf{0}, \sigma^2_t \textbf{I} \bigl),
    \label{eq:unconditional_langevin}
\end{equation}
where $\tau^{t-1}$ is the trajectory at the previous time step. $\alpha^t$ and $\gamma^t$ are learned coefficients that control the strength of the denoising process, while $\xi$ represents Gaussian noise.

\paragraph{Energy-based Models (EBM) with Composition.}
Energy-based models (EBMs)~\citep{hinton2002training,du2019implicit} are a class of generative models that define a probability distribution over data via an unnormalized energy function. For a given data $\tau \in \mathbb{R}^D $, the probability is given by 
$p_\theta(\tau) \propto e^{-E_\theta(\tau)},$ where $E_\theta(\tau) $ is a learnable neural network that parameterizes the energy. The sampling update rule is given by:
\begin{equation}
    \tau^{t-1}= \alpha^t(\tau^{t}-\gamma \nabla_{\tau} E_{\theta}(\tau^{t}) + \xi),\quad \xi\sim \mathcal{N} \bigl(\textbf{0}, \sigma^2_t \textbf{I} \bigl),
    \label{eq:unconditional_ebm}
\end{equation}
where $\tau^t $ represents the state (\ie, trajectory in this paper) at timestep $t $, and $\nabla_{\tau} E_{\theta}(\tau^t) $ is the gradient of the energy function at the current state. This sampling process is functionally similar to Eq.~\ref{eq:unconditional_langevin}.

In the scenario of multiple independent energy functions, the compositional EBMs~\citep{du2020compositional} are introduced, where the overall distribution is the product of several individual distributions: 
\begin{equation}
    p_{\text{product}}(\tau)  \propto p_\theta^1(\tau)p_\theta^2(\tau)\cdots p_\theta^N(\tau) \propto e^{- \sum_{i=1}^{n} E_{\theta}^i(\tau)  }.
    \label{eq:compose}
\end{equation}
The corresponding Langevin dynamics update rule for compositional EBMs is given by:
\begin{equation}
    \tau^{t-1}= \alpha^t\big(\tau^{t}-\gamma \big(\sum_{i=1}^{n}\nabla_{\tau} E_{\theta}(\tau^{t})\big) + \xi\big),\quad \xi\sim \mathcal{N} \bigl(\textbf{0}, \sigma^2_t \textbf{I} \bigl).
    \label{eq:compose_ebm}
\end{equation}
Our compositional diffusion policy draws inspiration from the principles of compositional EBMs, enabling the combination of multiple distribution-level outputs during the sampling process.

\begin{figure}[t]
    \centering
    \includegraphics[width = .8\linewidth]{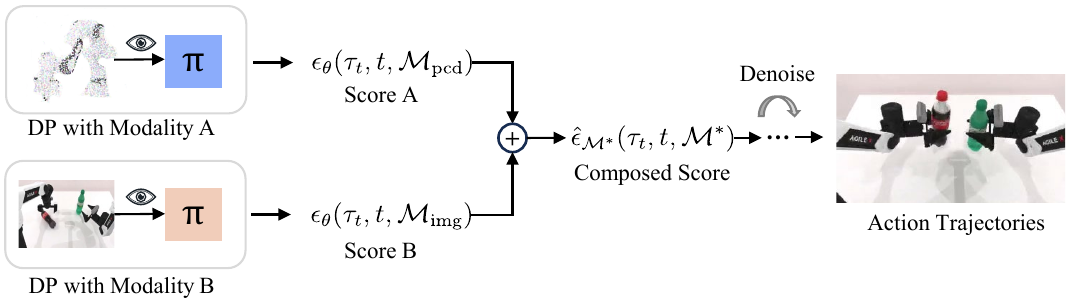}
    \caption{\textbf{Overview of our proposed modality-composable diffusion policy.} Combining distributional scores from pre-trained diffusion policies on different modalities (\ie, point cloud and image), MCDP can generate expressive and adaptable action trajectories without additional training.}
    \label{fig:pipeline}
    \vspace{-5mm}
\end{figure}

\section{Method}
In this section, we introduce the methodology for combining diffusion policies based on different single visual modalities. The core idea is to leverage compositional principles in the distribution level to combine score functions of diffusion effectively. First, we introduce the mathematical formulation of compositional diffusion models conditioning on different concepts in Sec.~\ref{subsec:cdm}. Building on this concept, we extend it to propose the modality-composable diffusion policy (MCDP) in Sec.~\ref{subsec:mcdp}, where the concept is specific as the single visual modality. The proposed framework allows for the combination of different modalities to enhance the flexibility and generalization of the learning process, enabling the MCDP to leverage multiple sources of visual information to improve task performance in complex, multimodal environments.

\subsection{Compositional Diffusion Models}
\label{subsec:cdm}
The key idea of the compositional diffusion model (CDM) is to model the distribution of a trajectory $\tau$ conditioned on multiple concepts $c_i$, similar to the compositional EBMs.
Mathematically, we can express the joint probability of the trajectory $\tau$ based on the set of concepts $\{c_1, \ldots, c_n\}$ in Eq.~\ref{eq:conj1}, and further reformulate the conditional terms by parameterizing $p(\vc_i|\tau) \propto \left(\frac{p(\tau|\vc_i)}{p(\tau)}\right)^\alpha$, as follows:
\begin{align}
   p(\tau|\vc_1, \ldots, \vc_n) & \propto p(\tau, \vc_1, \ldots, \vc_n) = p(\tau) \prod_{i=1}^n p(\vc_i|\tau), \label{eq:conj1}\\
   &  \propto p(\tau) \prod_{i=1}^n \left(\frac{p(\tau|\vc_i)}{p(\tau)}\right)^\alpha, \text{~with~} p(\vc_i|\tau) \propto \left(\frac{p(\tau|\vc_i)}{p(\tau)}\right)^\alpha, \label{eq:conj2}
\end{align}
where $p(\vc_i|\tau)$ can be interpreted as an implicit classifier~\citep{ho2022classifier} and $\alpha$ serves as a weighting factor that modulates the influence of each concept on the overall trajectory distribution.

Then, the score function of the composed distribution 
can be derived directly from Eq.~\ref{eq:conj2}:
\begin{align}
    \nabla_{\tau} \log p(\tau|\vc_1, \ldots, \vc_n) = \nabla_{\tau} \log p(\tau) + \sum_{i=1}^{n} \alpha\bigl(\nabla_{\tau}\log p(\tau|c_i) - \nabla_{\tau}\log p(\tau)\bigl).
\end{align}
Using the relationship between the score function of the distribution and diffusion score~\citep{baoanalytic}, \ie, $\epsilon_\theta(\tau_t, t) = - \sigma_\tau \nabla_\tau\log p(\tau)$, we can express the sampling process for the compositional diffusion model as follows:
\begin{align}
   \hat{\epsilon}(\tau_t, t, \vc) = \epsilon_\theta(\tau_t, t) + \sum_{i=1}^n w_i \bigl(\epsilon_\theta(\tau_t, t,  \vc_i) - \epsilon_{\theta}(\tau_t, t)\bigl), \label{eq:cfg}
\end{align}
where $\epsilon_\theta(\tau_t, t, \vc_i)/\epsilon_\theta(\tau_t, t)$ represents the noise estimation at time step $t $ for trajectory $\tau_t$ conditioned on the individual concept $\vc_i $ or without condition. The weights $w_i $ modulate the influence of each concept on the overall noise estimate.
This formulation represents a generalization of the classifier-free guidance (CFG)~\citep{ho2022classifier} technique commonly used in generative models. 

\subsection{Modality-Composable Diffusion Policy}
\label{subsec:mcdp}

Based on the previous foundation, we can now apply the CDM to diffusion policy for robotic tasks. Specifically, we aim to combine multiple diffusion policies, where each policy is conditioned on a different single visual modality, to improve the overall performance in multimodal settings (Fig.~\ref{fig:pipeline}).

The joint probability distribution of the trajectory conditioned on different modalities can be expressed as follows:
\begin{align}
   p_{\mathcal{M}^*}(\tau|\vc_1, \ldots, \vc_n) & \propto p_{\mathcal{M}_1}(\tau| \vc_1)p_{\mathcal{M}_2}(\tau| \vc_2)\cdots p_{\mathcal{M}_n}(\tau| \vc_n),
\end{align}

Similar to Eq.~\ref{eq:cfg}, we can apply the CFG sampling method to obtain a modality-composable diffusion policy. However, instead of using the CFG sampling, we opt for the original sampling process 
for the following reasons:
\textbf{(a) Policy flexibility:} The policies used for composition can either be trained individually or derived from existing open-source models. Notably, most current DP-based models, \eg, DP3~\citep{ze20243d}, Octo~\citep{team2024octo}, and RDT~\citep{liu2024rdt}, do not typically incorporate CFG training, making our CFG-free composition approach both practical and flexible.
\textbf{(b) Sampling efficiency:} CFG-based sampling requires performing a double computation to obtain both the unconditional and conditional score, which can significantly slow down the process. This efficiency loss is particularly critical in real-time robot control applications.

\begin{figure*}[t!] 
  \begin{minipage}[t]{0.5\textwidth} 
    \raggedleft
    \includegraphics[width=\linewidth]{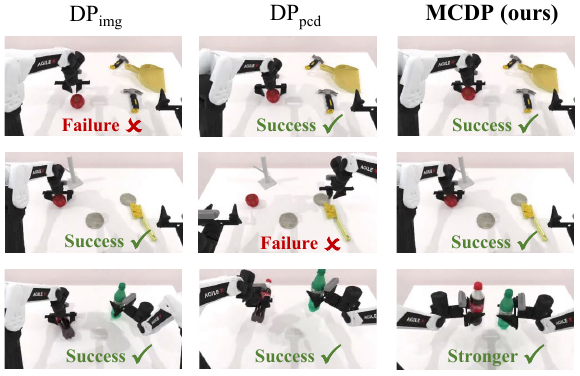} 
    \caption{\small \textbf{Visualization results of different diffusion policies.} Our proposed MCDP can be successful even when one part of the DP fails, and show stronger performance when both parts of the DP work.}
    \label{fig:vis_dps}
  \end{minipage} \hfill
  \begin{minipage}[t]{0.48\textwidth} 
  \vspace*{-120pt}
    \small
    \captionsetup{width=0.90\linewidth}
    \begin{tabular}{p{1\linewidth}} 
        \rule{\linewidth}{1pt}  
        \vspace{-0.3em}
        \noindent\textbf{ Alg.~1: Modality-Composable Diffusion Sampling} \\[-0.3em]
        \rule{\linewidth}{0.5pt}  
        \noindent\textbf{\hspace{1em} Input:} Pre-trained diffusion policies $\pi_1$, $\pi_2$, weights $w_1$, $w_2$, input modalities $\mathcal{M}_1$, $\mathcal{M}_2$  \\
        1: Initialize noise trajectory $\tau_N \sim \mathcal{N}(0, I)$ \\
        2: \textbf{for} $t = N, \dots, 1:$ \scriptsize\textcolor{gray}{// denoising steps} \\
        3: \hspace{1em} $\epsilon_1 \gets \pi_1(\tau_t, t, \mathcal{M}_1)$ \\
        4: \hspace{1em} $\epsilon_2 \gets \pi_2(\tau_t, t, \mathcal{M}_2)$ 
            \scriptsize \textcolor{gray}{  \# score estimation} \\
        5: \hspace{1em} $\epsilon_{\mathcal{M}^*} \gets w_1 * \epsilon_1 + w_2 * \epsilon_2$ 
            \scriptsize\textcolor{gray}{\# score composition} \\
        6: \hspace{1em} $\tau^{t-1}\gets\alpha^t(\tau^{t}-\gamma^t \epsilon_{\mathcal{M}^*} +\xi)$ \\
        \noindent\textbf{Return:} Action trajectory $\tau_0$ \\
        \vspace{-1em}
        \rule{\linewidth}{1pt}  
    \end{tabular}
    \vspace*{-5pt}
    \captionof{algorithm}{\small \textbf{Modality-Composable Diffusion Sampling.} During inference time, two policies are integrated through score composition to form a stronger MCDP.}
    \label{alg:psuedocode}
  \end{minipage}
  \vspace{-15pt}
\end{figure*}

As a result, our proposed MCDP (Alg.~\ref{alg:psuedocode}) does not rely on CFG, maintaining both general applicability and computational efficiency. The sampling process for our MCDP can be expressed as:
\begin{equation}
\hat{\epsilon}_{\mathcal{M}^*}(\tau_t, t, \vc) = \sum_{i=1}^{n} w_i \left( \epsilon_\theta(\tau_t, t,  \vc_i) \right), \quad \text{with} \quad \sum_{i=1}^{n} w_i = 1,
\end{equation}
where $ \epsilon_\theta(\tau_t, t, \vc_i) $ denotes the noise estimate conditioned on visual modality $\vc_i$ (\ie, $\mathcal{M}_i$), and $ w_i $ represents the weight assigned to each concept,
ensuring a balanced contribution from all modalities in the final trajectory estimate.
Since $ w_i $ is a parameter that needs to be manually tuned, we conduct extensive ablation experiments across different scenarios to provide empirical guidance on selecting optimal values for $ w_i $ in the following experiment section.

\section{Experiment}

We conduct experiments to investigate two key questions: 
1) How do different weight configurations influence the performance of Modality-Composable Diffusion Policy (MCDP) across various scenarios?
2) How can the advantages of the composed DP be explained?

\subsection{Influence of Weight Configurations on MCDP Performance}
To analyze the first question, we evaluate MCDP performance across multiple tasks in Tab.~\ref{tab:main_exp}. Several findings are summarized:

\includegraphics[height=.8em]{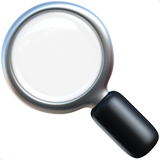}~\textbf{Finding 1: When both DPs have moderate accuracy (both $>$30\%), MCDP often achieves higher accuracy under appropriate weight configurations compared to unimodal DPs.}
For instance, in the Empty Cup Place task, DP$_{\text{img}}$ and DP$_{\text{pcd}}$ achieve 0.42 and 0.62, respectively, while MCDP peaks at 0.86 with $w_1 = 0.4$, surpassing both unimodal DPs. This improvement reflects the composition of diffusion scores capturing a more generalized distribution that reduces the reliance on specific modalities, consistent with the theoretical advantages of compositional models.

\includegraphics[height=.8em]{Figures/mag.png}~\textbf{Finding 2: When one DP has significantly lower accuracy, MCDP struggles to surpass the highest accuracy of the better-performing unimodal DP.}
For example, in the Pick Apple Messy task, DP$_{\text{pcd}}$ achieves 0.26, but MCDP peaks at 0.25 ($w_1 = 0.1$), falling short of DP$_{\text{pcd}}$. This suggests that low-accuracy scores from weaker modalities can significantly impact the joint distribution, diminishing the overall performance of the composed policy.

\includegraphics[height=.8em]{Figures/mag.png}~\textbf{Finding 3: Building on Finding 1, the improvement of MCDP is maximized when the better-performing unimodal DP holds a larger weight in MCDP.} 
For instance, in Dual Bottles Pick (Easy), where DP$_{\text{img}}$ achieves 0.77, MCDP reaches 0.85 with $w_1 = 0.8$, leveraging the stronger DP effectively. This highlights the necessity of assigning higher weights to the better-performing unimodal distribution to maximize the effectiveness of MCDP, offering a practical guideline for weight configuration in future applications.

These findings highlight the versatility of MCDP in leveraging the strengths of individual modalities and the importance of appropriately tuning the weights based on the performance characteristics of the unimodal DPs. 

\definecolor{light_green}{HTML}{E3F2D9}
\definecolor{deep_green}{HTML}{C8E5B3}
\definecolor{light_orange}{HTML}{fae5d3}
\definecolor{mid_orange}{HTML}{F5BEA5}
\definecolor{deep_orange}{HTML}{E77A61}

\begin{table}[t]
\centering
\renewcommand\arraystretch{1.2}
\caption{\textbf{Experiment results of our method under different composition configurations.}}
\vspace{-2mm}
\label{tab:main_exp}
\scalebox{0.72}{
\begin{tabular}{clcc|ccccccccc}
\toprule
\multirow{2}{*}{\textbf{Scenario}}& \multirow{2}{*}{{\textbf{Task}}} & \multirow{2}{*}{DP$_{\text{img}}$} & \multirow{2}{*}{DP$_{\text{pcd}}$} & \multicolumn{9}{c}{Modality-Composable Diffusion Policy}                                                           \\ \cline{5-13}
                                                 & &                      &                      & ${{0.1}^*}$ & $0.2$  & $0.3$ & $0.4$ & $0.5$ & $0.6 $ & $0.7$  & $0.8$  & $0.9$  \\ \hline
\multirow{3}{*}{\shortstack{\textit{Both Policies} \\  \textit{Perform Well}}} &Empty Cup Place  &0.42	&0.62   &0.70	&\cellcolor{deep_orange}0.86	&\cellcolor{mid_orange}0.84	&\cellcolor{deep_orange}\textbf{0.86}	&\cellcolor{mid_orange}0.84	&0.84	&0.76	&0.68	&0.61       \\
&Dual Bottles Pick (Hard)  &0.49	&0.64	&\cellcolor{mid_orange}0.69	&0.63	&\cellcolor{deep_orange}\textbf{0.71}	&0.66	&0.64	&0.65	&0.63	&0.56	&0.58 \\ 
&Shoe Place & 0.37 & 0.36 & 0.47 & 0.52	&0.56	&\cellcolor{mid_orange}0.59 &\cellcolor{deep_orange}0.60	&\cellcolor{mid_orange}0.59	&\cellcolor{mid_orange}0.59 & 0.53 & 0.41\\\hline
\multirow{2}{*}{\shortstack{ \textit{Both Policies} \\  \textit{Perform Bad}}} &Dual Shoes Place   &0.08	&\cellcolor{deep_orange}0.23		&0.19	&0.17	&0.19   & \cellcolor{mid_orange}0.20 &\cellcolor{mid_orange}0.20	&0.17	&0.16 &0.14	&0.09 \\ 
&Pick Apple Messy  &0.05	&\cellcolor{deep_orange}0.26		&\cellcolor{mid_orange}0.25	&0.17	&0.21	&0.15	&0.13	&0.08	&0.08	&0.06	&0.08  \\\hline
\multirow{1}{*}{\shortstack{{\textit{Policy A }} $>$ {\textit{Policy B}} }} & Dual Bottles Pick (Easy) &0.77	&0.36		&0.52	&0.64	&0.70	&0.75 &\cellcolor{mid_orange}0.82	&0.81	&0.80	&\cellcolor{deep_orange}0.85 & 0.80\\ \hline
\multirow{1}{*}{\shortstack{{\textit{Policy A }} $<$ {\textit{Policy B}} }} & Block Hammer Beat &0.00	&\cellcolor{deep_orange}0.76		&\cellcolor{mid_orange}0.61	&0.3	&0.18	&0.15	&0.12	&0.07	&0.00	&0.00	&0.00 \\\bottomrule
\multicolumn{13}{l}{
\begin{minipage}{1.3\textwidth}
    $^*$: The number set $\{0.1, ..., 0.9\}$ denotes the weight of DP$_{\text{img}}$ (\ie, $w_1$), 
    corresponding to the noise estimation of MCDP as $\hat{\epsilon}_{\mathcal{M}^*} = w_1 * \epsilon_{\text{DP}_{\text{img}}} + (1-w_1)*\epsilon_{\text{DP}_{\text{pcd}}}.$
\end{minipage}
}
\end{tabular}}
\vspace{-12pt}
\end{table}

\begin{figure}[t]
    \centering
    \includegraphics[width = 1.0\linewidth]{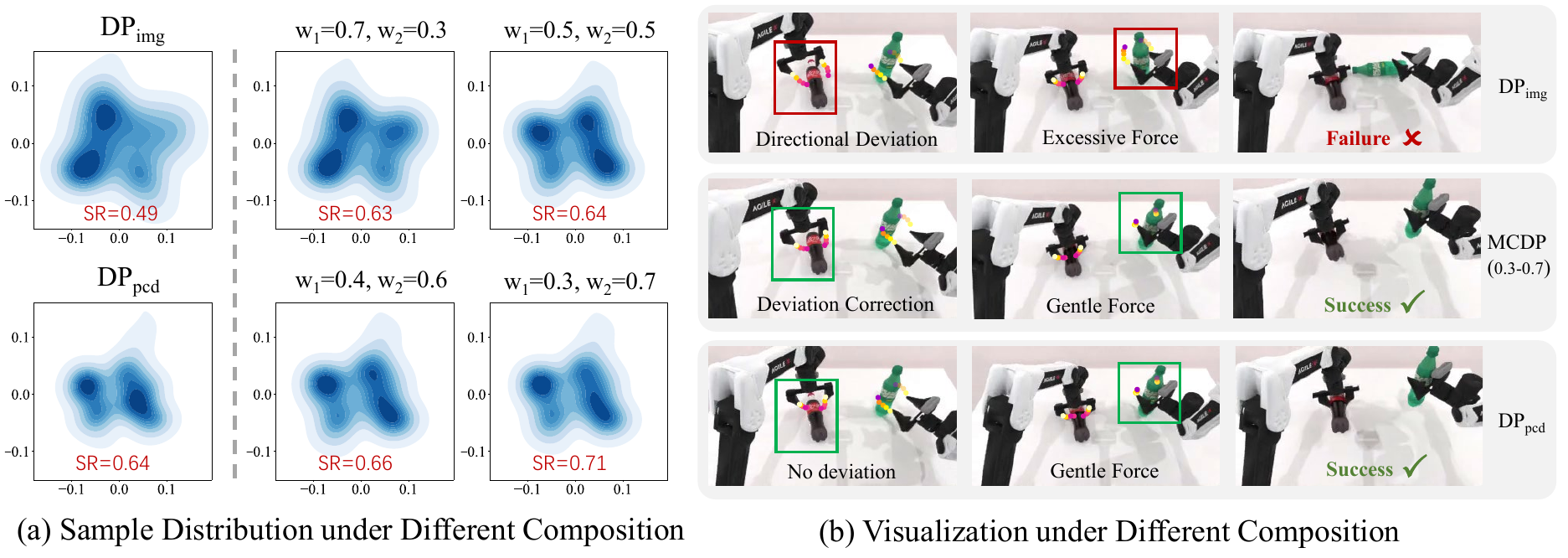}
    \caption{\textbf{Visual analysis of MCDP under different composition.}}
    \label{fig:dis_vis}
    \vspace{-15pt}
\end{figure}

\subsection{Analysis of MCDP's Advantage Based on Visualization}
For the second question, Fig.~\ref{fig:dis_vis}(a) illustrates the sample distribution of trajectories generated by MCDP under varying weight configurations. As the weights shift, the joint distribution adapts accordingly. For instance, when $w_1$ is larger, the distribution leans towards the characteristics of DP${_\text{img}}$, while larger $w_2$ emphasize DP$_{\text{pcd}}$. This visualization highlights how weight tuning refines the composed distribution to capture essential features from each modality, facilitating more successful policy execution.

The grasping process vividly demonstrates the advantage of MCDP over unimodal DPs in Fig.~\ref{fig:dis_vis}(b). DP$_{\text{img}}$, \eg, exhibits directional deviation and excessive force during execution, leading to task failure. On the other hand, MCDP effectively integrates the corrective trajectory from DP${_\text{pcd}}$, adjusting both direction and applied force to resolve the deviation and achieve successful grasping. This indicates MCDP's ability to mitigate errors from one modality by leveraging the strengths of another. 

\section{Conclusion}
In this work, we propose Modality-Composable Diffusion Policy (MCDP), a stronger diffusion-based policy that combines DPs through inference-time distribution composition. Extensive experiments show that MCDP improves adaptability and performance with the flexible composition of policies across different modalities. This work provides a promising direction for developing efficient and generalizable robot learning policies with the idea of composition. Our code is open-sourced at \url{https://github.com/AndyCao1125/MCDP}.

\bibliography{iclr2025_conference}

\begin{thebibliography}{35}
\providecommand{\natexlab}[1]{#1}
\providecommand{\url}[1]{\texttt{#1}}
\expandafter\ifx\csname urlstyle\endcsname\relax
  \providecommand{\doi}[1]{doi: #1}\else
  \providecommand{\doi}{doi: \begingroup \urlstyle{rm}\Url}\fi

\bibitem[Bao et~al.(2022)Bao, Li, Zhu, and Zhang]{baoanalytic}
Fan Bao, Chongxuan Li, Jun Zhu, and Bo~Zhang.
\newblock Analytic-dpm: an analytic estimate of the optimal reverse variance in diffusion probabilistic models.
\newblock In \emph{International Conference on Learning Representations (ICLR)}, 2022.

\bibitem[Black et~al.(2024)Black, Brown, Driess, Esmail, Equi, Finn, Fusai, Groom, Hausman, Ichter, et~al.]{black2024pi_0}
Kevin Black, Noah Brown, Danny Driess, Adnan Esmail, Michael Equi, Chelsea Finn, Niccolo Fusai, Lachy Groom, Karol Hausman, Brian Ichter, et~al.
\newblock $\backslash pi\_0 $: A vision-language-action flow model for general robot control.
\newblock \emph{arXiv preprint arXiv:2410.24164}, 2024.

\bibitem[Canonne(2020)]{canonne2020short}
Cl{\'e}ment~L Canonne.
\newblock A short note on learning discrete distributions.
\newblock \emph{arXiv preprint arXiv:2002.11457}, 2020.

\bibitem[Cao et~al.(2024)Cao, Zhang, Sun, Wang, Cheng, Li, Ma, Shao, Zhao, Han, et~al.]{cao2024mamba}
Jiahang Cao, Qiang Zhang, Jingkai Sun, Jiaxu Wang, Hao Cheng, Yulin Li, Jun Ma, Yecheng Shao, Wen Zhao, Gang Han, et~al.
\newblock Mamba policy: Towards efficient 3d diffusion policy with hybrid selective state models.
\newblock \emph{arXiv preprint arXiv:2409.07163}, 2024.

\bibitem[Chi et~al.(2023)Chi, Xu, Feng, Cousineau, Du, Burchfiel, Tedrake, and Song]{chi2023diffusion}
Cheng Chi, Zhenjia Xu, Siyuan Feng, Eric Cousineau, Yilun Du, Benjamin Burchfiel, Russ Tedrake, and Shuran Song.
\newblock Diffusion policy: Visuomotor policy learning via action diffusion.
\newblock \emph{The International Journal of Robotics Research (IJRR)}, pp.\  02783649241273668, 2023.

\bibitem[Du \& Kaelbling(2024)Du and Kaelbling]{du2024compositional}
Yilun Du and Leslie Kaelbling.
\newblock Compositional generative modeling: A single model is not all you need.
\newblock \emph{arXiv preprint arXiv:2402.01103}, 2024.

\bibitem[Du \& Mordatch(2019)Du and Mordatch]{du2019implicit}
Yilun Du and Igor Mordatch.
\newblock Implicit generation and modeling with energy based models.
\newblock \emph{Advances in Neural Information Processing Systems (NeurIPS)}, 32, 2019.

\bibitem[Du et~al.(2020)Du, Li, and Mordatch]{du2020compositional}
Yilun Du, Shuang Li, and Igor Mordatch.
\newblock Compositional visual generation with energy based models.
\newblock \emph{Advances in Neural Information Processing Systems (NeurIPS)}, 33:\penalty0 6637--6647, 2020.

\bibitem[Du et~al.(2023)Du, Li, Torralba, Tenenbaum, and Mordatch]{duimproving}
Yilun Du, Shuang Li, Antonio Torralba, Joshua~B Tenenbaum, and Igor Mordatch.
\newblock Improving factuality and reasoning in language models through multiagent debate.
\newblock In \emph{International Conference on Machine Learning (ICML)}, 2023.

\bibitem[Grathwohl et~al.(2020)Grathwohl, Wang, Jacobsen, Duvenaud, and Zemel]{grathwohl2020learning}
Will Grathwohl, Kuan-Chieh Wang, J{\"o}rn-Henrik Jacobsen, David Duvenaud, and Richard Zemel.
\newblock Learning the stein discrepancy for training and evaluating energy-based models without sampling.
\newblock In \emph{International Conference on Machine Learning (ICML)}, pp.\  3732--3747. PMLR, 2020.

\bibitem[Hinton(2002)]{hinton2002training}
Geoffrey~E Hinton.
\newblock Training products of experts by minimizing contrastive divergence.
\newblock \emph{Neural Computation}, 14\penalty0 (8):\penalty0 1771--1800, 2002.

\bibitem[Ho \& Salimans(2022)Ho and Salimans]{ho2022classifier}
Jonathan Ho and Tim Salimans.
\newblock Classifier-free diffusion guidance.
\newblock \emph{arXiv preprint arXiv:2207.12598}, 2022.

\bibitem[Ho et~al.(2020)Ho, Jain, and Abbeel]{ho2020denoising}
Jonathan Ho, Ajay Jain, and Pieter Abbeel.
\newblock Denoising diffusion probabilistic models.
\newblock \emph{Advances in Neural Information Processing Systems (NeurIPS)}, 33:\penalty0 6840--6851, 2020.

\bibitem[Hu et~al.(2024)Hu, Guo, Wang, Chen, Wang, Zhang, Sreenath, Lu, and Chen]{hu2024videopredictionpolicygeneralist}
Yucheng Hu, Yanjiang Guo, Pengchao Wang, Xiaoyu Chen, Yen-Jen Wang, Jianke Zhang, Koushil Sreenath, Chaochao Lu, and Jianyu Chen.
\newblock Video prediction policy: A generalist robot policy with predictive visual representations, 2024.
\newblock URL \url{https://arxiv.org/abs/2412.14803}.

\bibitem[Huang et~al.(2025)Huang, Chen, Zhou, Chen, Jiang, Hu, Gao, Li, Yao, and Ren]{huang2025enerverse}
Siyuan Huang, Liliang Chen, Pengfei Zhou, Shengcong Chen, Zhengkai Jiang, Yue Hu, Peng Gao, Hongsheng Li, Maoqing Yao, and Guanghui Ren.
\newblock Enerverse: Envisioning embodied future space for robotics manipulation.
\newblock \emph{arXiv preprint arXiv:2501.01895}, 2025.

\bibitem[Janner et~al.(2022)Janner, Du, Tenenbaum, and Levine]{janner2022planning}
Michael Janner, Yilun Du, Joshua Tenenbaum, and Sergey Levine.
\newblock Planning with diffusion for flexible behavior synthesis.
\newblock In \emph{International Conference on Machine Learning (ICML)}, pp.\  9902--9915. PMLR, 2022.

\bibitem[Karamcheti et~al.(2024)Karamcheti, Nair, Balakrishna, Liang, Kollar, and Sadigh]{karamchetiprismatic}
Siddharth Karamcheti, Suraj Nair, Ashwin Balakrishna, Percy Liang, Thomas Kollar, and Dorsa Sadigh.
\newblock Prismatic vlms: Investigating the design space of visually-conditioned language models.
\newblock In \emph{International Conference on Machine Learning (ICML)}, 2024.

\bibitem[Kim et~al.(2024)Kim, Pertsch, Karamcheti, Xiao, Balakrishna, Nair, Rafailov, Foster, Lam, Sanketi, et~al.]{kim2024openvla}
Moo~Jin Kim, Karl Pertsch, Siddharth Karamcheti, Ted Xiao, Ashwin Balakrishna, Suraj Nair, Rafael Rafailov, Ethan Foster, Grace Lam, Pannag Sanketi, et~al.
\newblock Openvla: An open-source vision-language-action model.
\newblock \emph{arXiv preprint arXiv:2406.09246}, 2024.

\bibitem[Liu et~al.(2021)Liu, Li, Du, Tenenbaum, and Torralba]{liu2021learning}
Nan Liu, Shuang Li, Yilun Du, Josh Tenenbaum, and Antonio Torralba.
\newblock Learning to compose visual relations.
\newblock \emph{Advances in Neural Information Processing Systems (NeurIPS)}, 34:\penalty0 23166--23178, 2021.

\bibitem[Liu et~al.(2024)Liu, Wu, Li, Tan, Chen, Wang, Xu, Su, and Zhu]{liu2024rdt}
Songming Liu, Lingxuan Wu, Bangguo Li, Hengkai Tan, Huayu Chen, Zhengyi Wang, Ke~Xu, Hang Su, and Jun Zhu.
\newblock Rdt-1b: a diffusion foundation model for bimanual manipulation.
\newblock \emph{arXiv preprint arXiv:2410.07864}, 2024.

\bibitem[Luo et~al.(2024)Luo, Sun, Tenenbaum, and Du]{luo2024potential}
Yunhao Luo, Chen Sun, Joshua~B Tenenbaum, and Yilun Du.
\newblock Potential based diffusion motion planning.
\newblock In \emph{International Conference on Machine Learning (ICML)}, 2024.

\bibitem[Mu et~al.(2024)Mu, Chen, Peng, Chen, Gao, Zou, Lin, Xie, and Luo]{mu2024robotwin}
Yao Mu, Tianxing Chen, Shijia Peng, Zanxin Chen, Zeyu Gao, Yude Zou, Lunkai Lin, Zhiqiang Xie, and Ping Luo.
\newblock Robotwin: Dual-arm robot benchmark with generative digital twins (early version).
\newblock \emph{arXiv preprint arXiv:2409.02920}, 2024.

\bibitem[Nichol \& Dhariwal(2021)Nichol and Dhariwal]{nichol2021improved}
Alexander~Quinn Nichol and Prafulla Dhariwal.
\newblock Improved denoising diffusion probabilistic models.
\newblock In \emph{International Conference on Machine Learning (ICML)}, pp.\  8162--8171. PMLR, 2021.

\bibitem[Reuss et~al.(2024)Reuss, Ya{\u{g}}murlu, Wenzel, and Lioutikov]{reuss2024multimodal}
Moritz Reuss, {\"O}mer~Erdin{\c{c}} Ya{\u{g}}murlu, Fabian Wenzel, and Rudolf Lioutikov.
\newblock Multimodal diffusion transformer: Learning versatile behavior from multimodal goals.
\newblock In \emph{First Workshop on Vision-Language Models for Navigation and Manipulation at ICRA 2024}, 2024.

\bibitem[Shi et~al.(2023)Shi, Sharma, Zhao, and Finn]{shi2023waypoint}
Lucy~Xiaoyang Shi, Archit Sharma, Tony~Z Zhao, and Chelsea Finn.
\newblock Waypoint-based imitation learning for robotic manipulation.
\newblock In \emph{Conference on Robot Learning (CoRL)}, pp.\  2195--2209. PMLR, 2023.

\bibitem[Song et~al.(2020{\natexlab{a}})Song, Meng, and Ermon]{songdenoising}
Jiaming Song, Chenlin Meng, and Stefano Ermon.
\newblock Denoising diffusion implicit models.
\newblock In \emph{International Conference on Learning Representations (ICLR)}, 2020{\natexlab{a}}.

\bibitem[Song et~al.(2020{\natexlab{b}})Song, Sohl-Dickstein, Kingma, Kumar, Ermon, and Poole]{song2020score}
Yang Song, Jascha Sohl-Dickstein, Diederik~P Kingma, Abhishek Kumar, Stefano Ermon, and Ben Poole.
\newblock Score-based generative modeling through stochastic differential equations.
\newblock \emph{arXiv preprint arXiv:2011.13456}, 2020{\natexlab{b}}.

\bibitem[Sridhar et~al.(2024)Sridhar, Shah, Glossop, and Levine]{sridhar2024nomad}
Ajay Sridhar, Dhruv Shah, Catherine Glossop, and Sergey Levine.
\newblock Nomad: Goal masked diffusion policies for navigation and exploration.
\newblock In \emph{IEEE International Conference on Robotics and Automation (ICRA)}, pp.\  63--70. IEEE, 2024.

\bibitem[Team et~al.(2024)Team, Ghosh, Walke, Pertsch, Black, Mees, Dasari, Hejna, Kreiman, Xu, et~al.]{team2024octo}
Octo~Model Team, Dibya Ghosh, Homer Walke, Karl Pertsch, Kevin Black, Oier Mees, Sudeep Dasari, Joey Hejna, Tobias Kreiman, Charles Xu, et~al.
\newblock Octo: An open-source generalist robot policy.
\newblock \emph{arXiv preprint arXiv:2405.12213}, 2024.

\bibitem[Wang et~al.(2024)Wang, Zhao, Du, Adelson, and Tedrake]{wang2024poco}
Lirui Wang, Jialiang Zhao, Yilun Du, Edward~H Adelson, and Russ Tedrake.
\newblock Poco: Policy composition from and for heterogeneous robot learning.
\newblock \emph{Robotics: Science and Systems (RSS)}, 2024.

\bibitem[Wu et~al.(2024)Wu, Hou, Liu, Che, Ju, Yang, Li, Zhao, Xu, Yang, et~al.]{wu2024robomind}
Kun Wu, Chengkai Hou, Jiaming Liu, Zhengping Che, Xiaozhu Ju, Zhuqin Yang, Meng Li, Yinuo Zhao, Zhiyuan Xu, Guang Yang, et~al.
\newblock Robomind: Benchmark on multi-embodiment intelligence normative data for robot manipulation.
\newblock \emph{arXiv preprint arXiv:2412.13877}, 2024.

\bibitem[Yang et~al.(2023)Yang, Mao, Du, Wu, Tenenbaum, Lozano-P{\'e}rez, and Kaelbling]{yang2023compositional}
Zhutian Yang, Jiayuan Mao, Yilun Du, Jiajun Wu, Joshua~B Tenenbaum, Tom{\'a}s Lozano-P{\'e}rez, and Leslie~Pack Kaelbling.
\newblock Compositional diffusion-based continuous constraint solvers.
\newblock In \emph{Conference on Robot Learning (CoRL)}, pp.\  3242--3265. PMLR, 2023.

\bibitem[Ze et~al.(2024)Ze, Zhang, Zhang, Hu, Wang, and Xu]{ze20243d}
Yanjie Ze, Gu~Zhang, Kangning Zhang, Chenyuan Hu, Muhan Wang, and Huazhe Xu.
\newblock 3d diffusion policy.
\newblock \emph{Robotics: Science and Systems (RSS)}, 2024.

\bibitem[Zhang et~al.(2024)Zhang, Tang, and Yan]{zhang2024versatile}
Gengyu Zhang, Hao Tang, and Yan Yan.
\newblock Versatile navigation under partial observability via value-guided diffusion policy.
\newblock In \emph{IEEE/CVF Conference on Computer Vision and Pattern Recognition (CVPR)}, pp.\  17943--17951, 2024.

\bibitem[Zhu et~al.(2024)Zhu, Zhu, Li, Wen, Xu, Liu, Cheng, Shen, Peng, Feng, et~al.]{zhu2024scaling}
Minjie Zhu, Yichen Zhu, Jinming Li, Junjie Wen, Zhiyuan Xu, Ning Liu, Ran Cheng, Chaomin Shen, Yaxin Peng, Feifei Feng, et~al.
\newblock Scaling diffusion policy in transformer to 1 billion parameters for robotic manipulation.
\newblock \emph{arXiv preprint arXiv:2409.14411}, 2024.

\end{thebibliography}
\bibliographystyle{iclr2025_conference}

\appendix

\section{Appendix}

Due to the paper limit, we supplement the sections of related work and experiment settings in the appendix as follows.

\section{Related Work}

\paragraph{Composable Generative Models} 
Composability refers to the ability to combine multiple components or distributions into a unified representation while preserving the properties of the individual elements. This principle is crucial in generative modeling, as it allows models to flexibly adapt to diverse tasks by reusing learned components. By enabling modularity, composable generative models facilitate visual generation, language reasoning, and manipulation across various domains:
Energy-based models (EBMs)~\citep{hinton2002training,du2019implicit,grathwohl2020learning} support compositionality by summing energies, allowing factor- and object-level combinations. \citet{du2020compositional} unified perspectives of compositionality for visual generation, enabling recursive combinations and continual learning. \citet{liu2021learning} further improved EBMs for scene generation by factorizing relational structures, achieving faithful generation and editing of complex scenes.
(b) Language Generation:
\citet{duimproving} combined outputs from individual language models using multi-agent debate, forming a coherent representation for robust language sequence generation.
(c) Trajectory Composition:
\citet{janner2022planning} applied compositionality to planning with diffusion-based methods, excelling in long-horizon decision-making. \citet{yang2023compositional} addressed continuous constraint satisfaction in robotic planning, while \citet{luo2024potential} optimized motion planning by learning potential fields, surpassing traditional approaches.
(d) Policy Composition:
Policy Composition (PoCo)~\citep{wang2024poco} combines data distributions using diffusion models to learn generalized manipulation skills. Building on PoCo, our work adopts a similar philosophy, using the visual modality as the specified condition and offering empirical insights and guidance through comprehensive benchmark analyses.

\paragraph{Diffusion Models in Robot Learning}

Due to their flexibility and representational power, diffusion models~\citep{ho2020denoising,songdenoising,nichol2021improved} have found extensive applications in robot learning. Specifically, diffusion models offer a novel way to represent policies, where action trajectories are denoised from noise to real trajectories and then interact with the environment.
The concept of Diffusion Policy (DP~\citep{chi2023diffusion} was first proposed to model action spaces using diffusion, significantly enhancing expressiveness compared to traditional explicit policies (e.g., behavior cloning) or implicit policies (e.g., EBMs). Since then, numerous advancements have been made: multimodal DP such as MDT~\citep{reuss2024multimodal}, trajectory extraction approaches like AWE~\citep{shi2023waypoint}, DP3~\citep{ze20243d} utilizing point cloud representations to achieve state-of-the-art performance, and vision-language-action models, \eg, Octo~\citep{team2024octo}, $\pi0$~\citep{black2024pi_0} and RDT~\citep{liu2024rdt}.
Diffusion models have also been leveraged for learning visual representations to improve scene understanding and subsequently enhance policy performance. For example, Enerverse~\citep{huang2025enerverse} employs video diffusion models to simultaneously learn 3D and 2D task-relevant video representations for downstream policies. Similarly, Video Prediction Policy~\citep{hu2024videopredictionpolicygeneralist} integrates predictive visual representations to generalize across tasks.
In this work, we adopt both DP and DP3 as single-visual-based diffusion policies for our composable diffusion policy.

\section{Experiment Settings}

The DP$_{\text{img}}$ and DP$_{\text{pcd}}$ correspond to the diffusion policy based on RGB images (\ie, DP~\citep{chi2023diffusion}) and point cloud (\ie, DP3~\citep{ze20243d}), respectively. We adopted the codebase from RoboTwin~\citep{mu2024robotwin}\footnote{\url{https://github.com/TianxingChen/RoboTwin}}, and reproduced the DP$_{\text{img}}$ and DP$_{\text{pcd}}$ (without using point cloud color) with random seed $0$. Since the diffusion scores from different policies are composed at each denoising step (Alg.~\ref{alg:psuedocode}), we replace the default DDIM scheduler with inference step $10$ of DP$_{\text{pcd}}$ to DDPM with $100$ inference steps, aligned with the DDPM scheduler of DP$_{\text{img}}$'s. Notably, the diffusion schedulers here are not strictly restricted to DDPM, but can be expanded to any solvers (\eg, Analytic-DPM and DPM-Solver). If different solvers are simultaneously adopted, the denoising process should be carefully tuned.

\end{document}